\documentclass[runningheads]{llncs}
\usepackage{times}
\usepackage{xcolor}
\usepackage{soul}
\usepackage[utf8]{inputenc}
\usepackage{amsmath}
\usepackage{multirow}
\usepackage{color}
\usepackage{subfigure}
\usepackage{url}
\usepackage{CJK}
\usepackage{amsfonts}
\usepackage{graphicx}  
\begin{document}
\title{Effective Character-augmented Word Embedding \\ for Machine Reading Comprehension}
%
%
\author{Zhuosheng Zhang$^{1,2}$, Yafang Huang$^{1,2}$, Pengfei Zhu$^{1,2,3}$, Hai Zhao$^{1,2,}$\thanks{Corresponding author. This paper was partially supported by
		National Key Research and Development Program of China (No. 2017YFB0304100),
		National Natural Science Foundation of China (No. 61672343 and No. 61733011),
		Key Project of National Society Science Foundation of China (No. 15-ZDA041),
		The Art and Science Interdisciplinary Funds of Shanghai Jiao Tong University (No. 14JCRZ04).}  
}
\institute{	$^{1}$Department of Computer Science and Engineering, Shanghai Jiao Tong University \\
	$^{2}$Key Laboratory of Shanghai Education Commission for Intelligent Interaction \\ and Cognitive Engineering, Shanghai Jiao Tong University, Shanghai, 200240, China\\
	$^{3}$School of Computer Science and Software Engineering, East China Normal University, China\\
	{\tt \{zhangzs, huangyafang\}@sjtu.edu.cn, 10152510190@stu.ecnu.edu.cn} \\{\tt zhaohai@cs.sjtu.edu.cn}}
\maketitle              
\begin{abstract}
	Machine reading comprehension is a task to model relationship between passage and query. In terms of deep learning framework, most of state-of-the-art models simply concatenate word and character level representations, which has been shown suboptimal for the concerned task. In this paper, we empirically explore different integration strategies of word and character embeddings and propose a character-augmented reader which attends character-level representation to augment word embedding with a short list to improve word representations, especially for rare words. Experimental results show that the proposed approach helps the baseline model significantly outperform state-of-the-art baselines on various public benchmarks. 
	
\end{abstract}

\section{Introduction}

\noindent Machine reading comprehension (MRC) is a challenging task which requires computers to read and understand documents to answer corresponding questions, it is indispensable for advanced context-sensitive dialogue and interactive systems \cite{Huang2018Moon,zhang2018DUA,Zhu2018lingke}. There are two main kinds of MRC, user-query types \cite{Joshi2017TriviaQA,rajpurkar2016squad} and cloze-style \cite{Cui2016Consensus,hermann2015teaching,hill2015goldilocks}. The major difference lies in that the answers for the former are usually a span of texts while the answers for the latter are words or phrases. 

Most of recent proposed deep learning models focus on sentence or paragraph level attention mechanism \cite{Cui2016Attention,Dhingra2017Gated,kadlec2016text,Seo2016Bidirectional,Wang2017Gated} instead of word representations. As the fundamental part in natural language processing tasks, word representation could seriously influence downstream MRC models (readers). Words could be represented as vectors using word-level or character-level embedding. For word embeddings, each word is mapped into low dimensional dense vectors directly from a lookup table. Character embeddings are usually obtained by applying neural networks on the character sequence of each word and the hidden states are used to form the representation. Intuitively, word-level representation is good at capturing wider context and dependencies between words but it could be hard to represent rare words or unknown words. In contrast, character embedding is more expressive to model sub-word morphologies, which facilitates dealing with rare words.

\begin{CJK*}{UTF8}{gkai}
	\begin{table*}
		\centering \scriptsize
				\caption{\label{tab:exp} A cloze-style reading comprehension example.}
		{
			\begin{tabular}{l|p{4.8cm}|p{6.1cm}}
				\hline
				\hline
				{\bf Passage }& 1 早上 ， 青蛙 、 小白兔 、 刺猬 和 大 蚂蚁 高高兴兴 过 桥 去 赶集 。
				
				2 不料 ， 中午 下 了 一 场 大暴雨 ， 哗啦啦 的 河水 把 桥 冲走 了 。
				
				3 天 快 黑 了 ， 小白兔 、 刺猬 和 大 蚂蚁 都 不 会 游泳 。
				
				4 过 不 了 河 ， 急 得 哭 了 。
				
				5 这时 ， 青蛙 想 ， 我 可 不 能 把 朋友 丢 下 ， 自己 过 河 回家 呀 。
				
				6 他 一面 劝 大家 不要 着急 ， 一 面 动脑筋 。
				
				7 嗬 ， 有 了 ！
				
				8 他 说 ： “ 我 有 个 朋 友 住 在 这儿 ， 我 去 找 他 想想 办法 。
				
				9 青蛙 找到 了 他 的 朋友 \underline{\hbox to 8mm{}}， 请求 他 说 ： “ 大家 过 不 了 河 了 ， 请 帮 个 忙 吧 ！
				
				10 鼹鼠 说 ： “ 可以 ， 请 把 大家 领到 我 家里 来 吧 。
				
				11 鼹鼠 把 大家 带 到 一个 洞口 ， 打开 了 电筒 ， 让 小白兔 、 刺猬 、 大 蚂蚁 和 青蛙 跟着 他 ， “ 大家 别 害怕 ， 一直 朝 前 走 。
				
				12 走 呀 走 呀 ， 只 听见 上面 “ 哗啦哗啦 ” 的 声音 ， 象 唱歌 。
				
				13 走 着 走 着 ， 突然 ， 大家 看见 了 天空 ， 天上 的 月
				亮 真 亮 呀 。
				
				14 小白兔 回头 一 瞧 ， 高兴 极了 ： “ 哈 ， 咱们 过 了 河 啦 ！
				
				15 唷 ， 真 了不起 。
				
				16 原来 ， 鼹鼠 在 河 底 挖 了 一 条 很 长 的 
				地道 ， 从 这 头 到 那 头 。
				
				17 青蛙 、 小白兔 、 刺猬 和 大 蚂蚁 是 多么 感激 鼹鼠 啊 ！
				
				18 第二 天 ， 青蛙 、 小白兔 、 刺猬 和 大 蚂蚁 带来 很多 很多 同伴 ， 杠 着 木头 ， 抬 着 石头 ， 要求 鼹鼠 让 他们 来 把 地道 挖 大 些 ， 修 成 河 底 大 “ 桥 ” 。
				
				19 不久 ， 他们 就 把 鼹鼠家 的 地道 ， 挖 成 了 河 底 的 一 条 大 隧道 ， 大家 可以 从 河 底 过 何 ， 还 能 通车 ， 真 有 劲 哩 !
				& 1 In the morning, the frog, the little white rabbit, the hedgehog and the big ant happily crossed the bridge for the market.
				
				2 Unexpectedly, a heavy rain fell at noon, and the water swept away the bridge.
				
				3 It was going dark. The little white rabbit, hedgehog and big ant cannot swim.
				
				4 Unable to cross the river, they were about to cry.
				
				5 At that time, the frog made his mind that he could not leave his friend behind and went home alone.
				
				6 Letting his friends take it easy, he thought and thought.
				
				7 Well, there you go!
				
				8 He said, “I have a friend who lives here, and I'll go and find him for help.”
				
				9 The frog found his friend \underline{\hbox to 8mm{}}  and told him, “We cannot get across the river. Please give us a hand!”
				
				10 The mole said, "That's fine, please bring them to my house."
				
				11 The mole took everyone to a hole, turned on the flashlight and asked the little white rabbit, the hedgehog, the big ant and the frog to follow him, saying, "Don't be afraid, just go ahead."
				
				12 They walked along, hearing the “walla-walla” sound, just like a song.
				
				13 All of a sudden, everyone saw the sky, and the moon was really bright.
				
				14 The little white rabbit looked back and rejoiced: “ha, the river crossed!”.
				
				15 “Oh, really great.”
				
				16 Originally, the mole dug a very long tunnel under the river, from one end to the other.
				
				17 How grateful the frog, the little white rabbit, the hedgehog and the big ant felt to the mole!
				
				18 The next day, the frog, the little white rabbit, the hedgehog, and the big ant with a lot of his fellows, took woods and stones. They asked the mole to dig tunnels bigger, and build a great bridge under the river.
				
				19 It was not long before they dug a big tunnel under the river, and they could pass the river from the bottom of the river, and it could be open to traffic. It is amazing!\\
				\hline
				{\bf Query} & 青蛙 找到 了 他 的 朋友 \underline{\hbox to 8mm{}} ， 请求 他 说 ： “ 大家 过 不 了 河 了 ， 请 帮 个 忙 吧 ！ ” & The frog found his friend \underline{\hbox to 8mm{}} and told him, “We cannot get across the river. Please give us a hand!” \\
				\hline
				{\bf Answer} & 鼹鼠 & the mole \\
				\hline
				\hline
			\end{tabular}
		}

	\end{table*}
\end{CJK*}

As shown in Table \ref{tab:exp}, the passages in MRC are quite long and diverse which makes it hard to record all the words in the model vocabulary. As a result, reading comprehension systems suffer from out-of-vocabulary (OOV) word issues, especially when the ground-truth answers tend to include rare words or named entities (NE) in cloze-style MRC tasks. 

To form a fine-grained embedding, there have been a few hybrid methods that jointly learn the word and character representations \cite{Kim2015Character,Yang2016Words,luong2016achieving}. However, the passages in machine reading dataset are content-rich and contain massive words and characters, using fine-grained features, such as named entity recognition and part-of-speech (POS) tags will need too high computational cost in return, meanwhile the efficiency of readers is crucial in practice. 

In this paper, we verify the effectiveness of various simple yet effective character-augmented word embedding (CAW) strategies and propose a CAW Reader. We survey different CAW strategies to integrate word-level and character-level embedding for a fine-grained word representation. To ensure adequate training of OOV and low-frequency words, we employ a short list mechanism. Our evaluation will be performed on three public Chinese reading comprehension datasets and one English benchmark dataset for showing our method is effective in multi-lingual case.

\section{Related Work}

Machine reading comprehension has been witnessed rapid progress in recent years \cite{sordoni2016iterative,Trischler2016Natural,Wang2016Machine,Munkhdalai2016Reasoning,Wang2017Conditional,Dhingra2017Gated,zhang2018SubMRC,zhang2018OneShot,Yizhong2018Multi}. Thanks to various released MRC datasets, we can evaluate MRC models in different languages. This work focuses on cloze-style ones since the answers are single words or phrases instead of text spans, which could be error-prone when they turn out to be rare or OOV words that are not recorded in the model vocabulary.

Recent advances for MRC could be mainly attributed to attention mechanisms, including query-to-passage attention \cite{kadlec2016text,Cui2016Consensus}, attention-over-attention \cite{Cui2016Attention} and self attention \cite{Wang2017Gated}. Different varieties and combinations have been proposed for further improvements \cite{Dhingra2017Gated,Seo2016Bidirectional}. However, the fundamental part, word representation, which proves to be quite important in this paper, has not aroused much interest. To integrate the advantages of both word-level and character-level embeddings, some researchers studied joint models for richer representation learning where the common combination method is the concatenation. Seo et al. \cite{Seo2016Bidirectional} concatenated the character and word embedding and then fed the joint representation to a two-layer Highway Network. FG reader in  \cite{Yang2016Words} used a fine-grained gating mechanism to dynamically combine word-level and character-level representations based on word property. However, this method is computationally complex and requires extra labels such as NE and POS tags. 

Not only for machine reading comprehension tasks, character embedding has also benefited other natural language process tasks, such as word segmentation \cite{Cai2016Neural,Cai2017Fast}, machine translation \cite{Ling2015Character,luong2016achieving}, tagging \cite{lample2016neural,Bai2018deep,He2018Syntax} and language modeling \cite{Miyamoto2016Gated,Peters2018Deep}. Notablely, Cai et al. \cite{Cai2017Fast} presented a greedy neural word segmenter where high-frequency word embeddings are attached to character embedding via average pooling while low-frequency words are represented as character embedding. Experiments show this mechanism helps achieve state-of-the-art word segmentation performance, which partially inspires our reader design.

\section{Model}

In this section, we will introduce our model architecture, which is consisted of a fundamental word representation module and a gated attention learning module. 

\begin{figure*}
	\centering
	\includegraphics[width=0.6\textwidth]{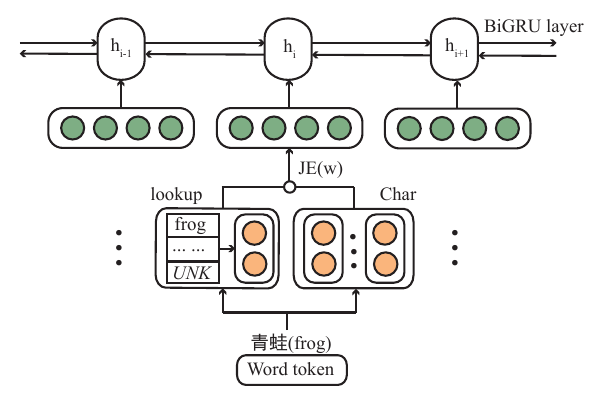}
	\caption{Overview of the word representation module.}
	\label{fig:framework}
\end{figure*}

\subsection{Word Representation Module}
Figure \ref{fig:framework} illustrates our word representation module. The input token sequence is first encoded into embeddings. In the context of machine reading comprehension tasks, word only representation generalizes poorly due to the severe word sparsity, especially for rare words. We adopt two methods to augment word representations, namely, a short list filtering and character enhancement. 

Actually, if all the words in the dataset are used to build the vocabulary, the OOV words from the test set will not be well dealt with for inadequate training. To handle this issue, we keep a short list $L$ for specific words. If word $w$ is in $L$, the immediate word embedding $\mathbf{e}_w$ is indexed from word lookup table $M^{w} \in \mathbb{R}^{d \times s}$ where $s$ denotes the size (recorded words) of lookup table and $d$ denotes the embedding dimension. Otherwise, it will be represented as the randomly initialized default word (denoted by a specific mark $UNK$). Note that only the word embedding of the OOV words will be replaced by the vectors of $UNK$ (denoted by $\mathbf{e}_u$) while their character embedding $\mathbf{e}_c$ will still be processed using the original word. In this way, the OOV words could be tuned sufficiently with expressive meaning after training.

In our experiments, the short list is determined according to the word frequency. Concretely, we sort the vocabulary according to the word frequency from high to low. A frequency filter ratio $\gamma$ is set to filter out the low-frequency words (rare words) from the lookup table. For example, $\gamma$=0.9 means the least frequent 10\% words are replaced with the default UNK notation.

Character-level embeddings have been widely used in lots of natural language processing tasks and verified for the OOV and rare word representations. Thus, we consider employing neural networks to compose word representations from smaller units, i.e., character embedding \cite{Miyamoto2016Gated,Kim2015Character}, which results in a hybrid mechanism for word representation with a better fine-grained consideration. For a given word $w$, a joint embedding (JE) is to straightforwardly integrate word embedding $\mathbf{e}_w$ and character embedding $\mathbf{e}_c$. 
\begin{align*}	
JE(w)= \mathbf{e}_w \circ  \mathbf{e}_c
\end{align*}
where $\circ$ denotes the joint operation. Specifically, we investigate concatenation (\emph{concat}), element-wise summation (\emph{sum}) and element-wise multiplication (\emph{mul}). Thus, each passage ${P}$ and query $Q$ is represented as $\mathbb{R}^{d \times k}$ matrix where $d$ denotes the dimension of word embedding and $k$ is the number of words in the input. 

Finally by combining the short list mechanism and character enhancement, $JE$($w$) can be rewritten as
\begin{align*}	
JE(w)=\left\{
\begin{array}{ll}
\mathbf{e}_w \circ \mathbf{e}_c   &   \text{if} \ w \in L\\
\mathbf{e}_u \circ  \mathbf{e}_c & \text{otherwise}\\
\end{array} \right.
\end{align*}

The character embedding $e_c$ can be learned by two kinds of networks, recurrent neural network (RNN) or convolutional neural network (CNN)\footnote{Empirical study shows the character embeddings obtained from these two networks perform comparatively. To focus on the performance of character embedding, we introduce the networks only for reproduction. Our reported results are based on RNN based character embeddings.}. 

\subsubsection{RNN based embedding}
The character embedding $\mathbf{e}_c$ is generated by taking the final outputs of a bidirectional gated recurrent unit (GRU) \cite{Cho2014Learning} applied to the vectors from a lookup table of characters in both forward and backward directions. Characters $w = \{x_{1},x_{2},\dots,x_{l}\}$ of each word are vectorized and successively fed to forward GRU and backward GRU to obtain the internal features. The output for each input is the concatenation of the two vectors from both directions: $\overleftrightarrow{h_{t}} = \overrightarrow{h_{t}}\parallel \overleftarrow{h_{t}}$ where $h_{t}$ denotes the hidden states.

Then, the output of BiGRUs is passed to a fully connected layer to obtain the a fixed-size vector for each word and we have $\mathbf{e}_c = W \overleftrightarrow{h_{t}} + b$.

\subsubsection{CNN based embedding}

character sequence $w = \{x_{1},x_{2},\dots,x_{l}\}$ is embedded into vectors $M$ using a lookup table, which is taken as the inputs to the CNN, and whose size is the input channel size of the CNN. Let $W_j$ denote the Filter matrices of width $l$, the substring vectors will be transformed to sequences $c_j (j \in [1, l])$:
\begin{align*}
c_j  = [\dots; \tanh (W_j\cdot M_{[i:i+l-1]} + b_j); \dots]
\end{align*}
where $[i:i+l-1]$ indexes the convolution window. A \emph{one-max-pooling} operation is adopted after convolution $s_{j} = \mathbf{max}(c_{j}) $. The character embedding is obtained through concatenating all the mappings for those $l$ filters.
\begin{align*}
\mathbf{e}_c &= [s_{1}\oplus \cdots  \oplus s_{j} \oplus \cdots \oplus s_{l}]
\end{align*}

\subsection{Attention Learning Module}
To obtain the predicted answer, we first apply recurrent neural networks to encode the passage and query. Concretely, we use BiGRUs to get contextual representations of forward and backward directions for each word in the passage and query and we have $G_{p}$ and $G_{q}$, respectively.

Then we calculate the gated attention following \cite{Dhingra2017Gated} to obtain the probability distribution of each word in the passage. For each word $p_{i}$ in $G_{p}$, we apply soft attention to form a word-specific representation of the query $q_{i} \in G_{q}$, and then multiply the query representation with the passage word representation.
\begin{align*}
\alpha_{i} &= softmax(G_{q}^{\top}p_{i}) \\
\beta_{i} &= G_{q}\alpha_{i} \\
x_{i} &= p_{i}\odot \beta_{i} \\
\end{align*}
where $\odot$ denotes the element-wise product to model the interactions between $p_{i}$ and $q_{i}$. The passage contextual representation ${\tilde G}_{p} = \{x_{1}, x_{2},\dots, x_{k}\}$ is weighted by query representation.

Inspired by \cite{Dhingra2017Gated}, multi-layered attentive network tends to focus on different aspects in the query and could help combine
distinct pieces of information to answer the query, we use $K$ intermediate layers which stacks end to end to learn the attentive representations. At each layer, the passage contextual representation ${\tilde G}_{p}$ is updated through above attention learning. Let $q_{k}$ denote the $k$-th intermediate output of query contextual representation and $G_{P}$ represent the full output of passage contextual representation $\tilde{G}_{p}$. The probability of each word $w \in C$ in the passage as being the answer is predicted using a softmax layer over the inner-product between $q_{k}$ and $G_{P}$.
\begin{align*}
r = softmax ((q_k)^{\top} G_{P})
\end{align*}
where vector $p$ denotes the probability distribution over all the words in the passage.
Note that each word may occur several times in the passage. Thus, the probabilities of each candidate word occurring in different positions of the passage are added together for final prediction.
\begin{align*}
P(w|p,q) \propto \sum_{i \in I(w,p)}r_{i}
\end{align*}
where $I(w,p)$ denotes the set of positions that a particular word $w$ occurs in the passage $p$. The training objective is to maximize $\log P(A|p,q)$ where $A$ is the correct answer. 

Finally, the candidate word with the highest probability will be chosen as the predicted answer. Unlike recent work employing complex attention mechanisms, our attention mechanism is much more simple with comparable performance so that we can focus on the effectiveness of our embedding strategies.
\section{Evaluation}
\subsection{Dataset and Settings}

\begin{table*}
	\centering
	\caption{\label{tab:dataset} Data statistics of PD, CFT and CMRC-2017. }
	{
		\begin{tabular}{cccccccc}
			\hline
			\hline
			& \multicolumn{3}{c}{PD}  & CFT & \multicolumn{3}{c}{CMRC-2017}  \\
			& Train & Valid & Test & human &Train & Valid & Test  \\
			\hline
			\# Query  &870,710 & 3,000 & 3,000  & 1,953& 354,295 & 2,000 & 3,000\\
			Avg \# words in docs  & 379 & 425 & 410 &   153 & 324 & 321  & 307\\
			Avg \# words in query & 38 & 38 & 41 & 20 & 27 & 19 & 23 \\
			\# Vocabulary  & \multicolumn{4}{c}{248,160} & \multicolumn{3}{c}{94,352} \\
			\hline
			\hline
		\end{tabular}
	}
	
\end{table*}

Based on three Chinese MRC datasets, namely People's Daily (PD), Children Fairy Tales (CFT) \cite{Cui2016Consensus} and CMRC-2017 \cite{Cui2017Dataset}, we verify the effectiveness of our model through a series of experiments\footnote{In the test set of CMRC-2017 and human evaluation test set (Test-human) of CFT, questions are further processed by human and the pattern of them may not be in accordance with the auto-generated questions, so it may be harder for machine to answer.}. Every dataset contains three parts, \emph{Passage}, \emph{Query} and \emph{Answer}. The \emph{Passage} is a story formed by multiple sentences, and the \emph{Query} is one sentence selected by human or machine, of which one word is replaced by a placeholder, and the the \emph{Answer} is exactly the original word to be filled in. The data statistics is shown in Table \ref{tab:dataset}. The difference between the three Chinese datasets and the current cloze-style English MRC datasets including Daily Mail, CBT and CNN \cite{hermann2015teaching} is that the former does not provide candidate answers. For the sake of simplicity, words from the whole passage are considered as candidates.

Besides, for the test of generalization ability in multi-lingual case, we use the Children's Book Test (CBT) dataset \cite{hill2015goldilocks}. We only consider cases of which the answer is either a NE or common noun (CN). These two subsets are more challenging because the answers may be rare words.  

For fare comparisons, we use the same model setting in this paper. We randomly initialize the 100$d$ character embeddings with the uniformed distribution in the interval [-0:05, 0:05]. We use word2vec \cite{mikolov:2013} toolkit to pre-train 200$d$ word embeddings on \emph{Wikipedia} corpus\footnote{\url{https://dumps.wikimedia.org/} }, and randomly initialize the OOV words. For both the word and character representation, the GRU hidden units are 128. For optimization, we use stochastic gradient descent with ADAM updates \cite{kingma2014adam}. The initial learning rate is 0.001, and after the second epoch, it is halved every epoch. The batch size is 64. To stabilize GRU training, we use gradient clipping with a threshold of 10. Throughout all experiments, we use three attention layers.

\subsection{Results}

\begin{table}
	\centering 
	\caption{\label{tab:pdcftresult} Accuracy on PD and CFT datasets. All the results except ours are from \cite{Cui2016Consensus}.}
	{
		\begin{tabular}{l|c|c|c|c}
			\hline
			\hline
			\multirow{2}{*}{Model} &\multirow{2}{*}{Strategy} & \multicolumn{2}{c}{PD}  & CFT  \\
			& & Valid & Test &Test-human  \\
			
			\hline
			AS Reader & - &  64.1 & 67.2 & 33.1  \\
			GA Reader & - &  64.1 & 65.2 & 35.7  \\
			CAS Reader & - &  65.2 & 68.1 & 35.0  \\
			\hline
			\multirow{3}{*}{CAW Reader} &concat  & 64.2 & 65.3 & 37.2    \\
			& sum  & 65.0 & 68.1 & 38.7    \\
			& mul & \textbf{69.4} & \textbf{70.5} & \textbf{39.7}    \\
			\hline
			\hline
		\end{tabular}
	}
	
\end{table}

\subsubsection{PD \& CFT}  

Table \ref{tab:pdcftresult} shows the results on PD and CFT datasets. With improvements of 2.4\% on PD and 4.7\% on CFT datasets respectively, our CAW Reader model significantly outperforms the CAS Reader in all types of testing.
Since the CFT dataset contains no training set, we use PD training set to train the corresponding model. It is harder for machine to answer because the test set of CFT dataset is further processed by human experts, and the pattern quite differs from PD dataset. We can learn from the results that our model works effectively for out-of-domain learning, although PD and CFT datasets belong to quite different domains.

\begin{table}
	\centering
	\caption{\label{tab:cmrc} Accuracy on CMRC-2017 dataset. Results marked with $\dag$ are from the latest official CMRC Leaderboard \protect\footnotemark. The best results are in bold face. WE is short for word embedding.  }
	{
		\begin{tabular}{l|c|c}
			\hline
			\hline
			\multirow{2}{*}{Model}  & \multicolumn{2}{c}{CMRC-2017} \\
			
			& Valid & Test \\
			\hline
			Random Guess \dag & 1.65 & 1.67\\
			Top Frequency \dag  & 14.85 & 14.07  \\
			AS Reader \dag & 69.75 & 71.23 \\
			GA Reader &  72.90 & 74.10 \\
			\cline{1-3}
			SJTU BCMI-NLP \dag & 76.15 & 77.73 \\
			6ESTATES PTE LTD \dag & 75.85 & 74.73  \\
			Xinktech \dag & 77.15 & 77.53 \\
			Ludong University \dag & 74.75 & 75.07 \\
			ECNU \dag & 77.95 & 77.40 \\
			WHU \dag & \textbf{78.20} & 76.53 \\
			
			\cline{1-3}
			CAW Reader (WE only) & 69.70 & 70.13 \\
			\cline{1-3}
			CAW Reader (concat)  & 71.55 & 72.03     \\
			CAW Reader (sum)  & 72.90 & 74.07      \\
			CAW Reader (mul) & 77.95  & \textbf{78.50}\\
			
			\hline
			\hline
		\end{tabular}
	}
\end{table}
\footnotetext{\url{http://www.hfl-tek.com/cmrc2017/leaderboard.html}}
\subsubsection{CMRC-2017} Table \ref{tab:cmrc} shows the results \footnote{Note that the test set of CMRC-2017 and human evaluation test set (Test-human) of CFT are  harder for the machine to answer because the questions are further processed manually and may not be in accordance with the pattern of auto-generated questions.}. Our CAW Reader (mul) not only obtains 7.27\% improvements compared with the baseline Attention Sum Reader (AS Reader) on the test set, but also outperforms all other single models. The best result on the valid set is from WHU, but their result on test set is lower than ours by 1.97\%, indicating our model has a satisfactory generalization ability.

We also compare different CAW strategies for word and character embeddings. We can see from the results that the CAW Reader (mul) significantly outperforms all the other three cases, word embedding only, concatenation and summation, and especially obtains 8.37\% gains over the first one. This reveals that compared with concatenation and sum operation, the element-wise multiplication might be more informative, because it introduces a similar mechanism to endow character-aware \emph{attention} over the word embedding. On the other hand, too high dimension caused by concatenation operation may lead to serious over-fitting issues \footnote{For the best concat and mul model, the training/validation accuracies are 97.66\%/71.55, 96.88\%/77.95\%, respectively.}, and sum operation is too simple to prevent from detailed information losing.

\subsubsection{CBT} The results on CBT are shown in Table \ref{tab:cbt}. Our model outperforms most of the previous public works. Compared with GA Reader with word and character embedding concatenation, i.e., the original model of our CAW Reader, our model with the character augmented word embedding has 2.4\% gains on the CBT-NE test set. FG Reader adopts neural gates to combine word-level and character-level representations and adds extra features including NE, POS and word frequency, but our model also achieves comparable performance with it. This results on both languages show that our CAW Reader is not limited to dealing with Chinese but also for other languages.

\begin{table}
	\centering
	\caption{\label{tab:cbt} Accuracy on CBT dataset. Results marked with $\ddag$ are of previously published works \cite{Dhingra2017Gated,Cui2016Consensus,Yang2016Words}.}
	{
		\begin{tabular}{l|c|c|c|c}
			\hline
			\hline
			\multirow{2}{*}{Model}  & \multicolumn{2}{c}{CBT-NE}  & \multicolumn{2}{c}{CBT-CN} \\
			& Valid & Test   & Valid & Test  \\
			\hline
			Human \ddag &- & 81.6 & - & 81.6 \\
			\hline
			LSTMs \ddag & 51.2 & 41.8 & 62.6 & 56.0 \\ 
			MemNets \ddag & 70.4 & 66.6 & 64.2 & 63.0 \\
			AS Reader \ddag & 73.8 & 68.6 & 68.8 & 63.4 \\
			Iterative Attentive Reader  \ddag & 75.2 & 68.2 & 72.1 & 69.2 \\
			EpiReader \ddag & 75.3 & 69.7 & 71.5 & 67.4 \\
			AoA Reader \ddag & 77.8 & 72.0 & 72.2 & 69.4 \\
			NSE \ddag & \textbf{78.2} & \textbf{73.2} & \textbf{74.3} & \textbf{71.9} \\
			\hline
			GA Reader \ddag & 74.9 & 69.0 & 69.0 & 63.9 \\
			GA word char concat \ddag & 76.8 & 72.5 & 73.1 & 69.6 \\
			GA scalar gate \ddag & 78.1 & 72.6 & 72.4 & 69.1 \\			
			GA fine-grained gate \ddag & 78.9 & 74.6 & 72.3 & 70.8 \\
			FG Reader \ddag & \textbf{79.1} & \textbf{75.0} & \textbf{75.3} & \textbf{72.0} \\	
			\hline
			CAW Reader& 78.4 & 74.9 & 74.8 & 71.5\\
			\hline
			\hline
		\end{tabular}
	}
	
\end{table}

\section{Analysis}

\begin{figure}
	\centering
	\includegraphics[width=0.85\textwidth]{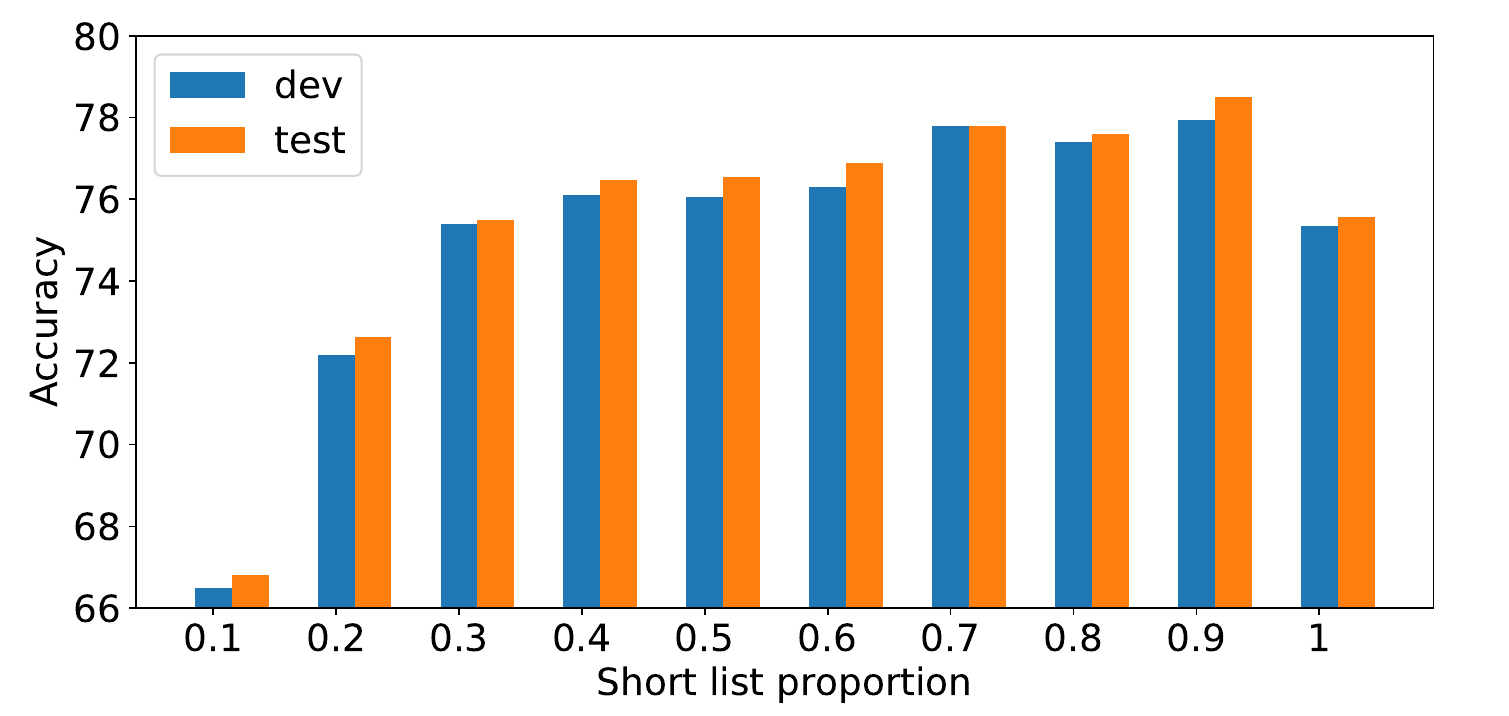}
	\caption{Quantitative study on the influence of the short list.}
	\label{fig:proportion}
	
\end{figure}

We conduct quantitative study to investigate how the short list influence the model performance on the filter ratio from [0.1, 0.2, \dots, 1]. Figure \ref{fig:proportion} shows the results on the CMRC-2017 dataset. Our CAW reader achieves the best accuracy when $\gamma=0.9$. It indicates that it is not optimal to build the vocabulary among the whole training set, and we can reduce the frequency filter ratio properly to promote the accuracy. In fact, training the model on the whole vocabulary may lead to over-fitting problems. Besides, improper initialization of the rare words may also bias the whole word representations. As a result, without a proper OOV representation mechanism, it is hard for a model to deal with OOV words from test sets precisely. 

\section{Conclusion}

This paper surveys multiple embedding enhancement strategies and proposes an effective embedding architecture by attending character representations to word embedding with a short list to enhance the simple baseline for the reading comprehension task. Our evaluations show that the intensified embeddings can help our model achieve state-of-the-art performance on multiple large-scale benchmark datasets. Different from most existing works that focus on either complex attention architectures or manual features, our model is more simple but effective. Though this paper is limited to the empirical verification on MRC tasks, we believe that the improved word representation may also benefit other tasks as well.
%
%
%
\bibliographystyle{splncs04}
%
\bibliography{mrc}

\begin{thebibliography}{10}
\providecommand{\url}[1]{\texttt{#1}}
\providecommand{\urlprefix}{URL }
\providecommand{\doi}[1]{https://doi.org/#1}

\bibitem{Bai2018deep}
Bai, H., Zhao, H.: Deep enhanced representation for implicit discourse relation
  recognition. In: Proceedings of the 27th International Conference on
  Computational Linguistics (COLING 2018) (2018)

\bibitem{Cai2016Neural}
Cai, D., Zhao, H.: Neural word segmentation learning for {Chinese}. In:
  Proceedings of the 54th Annual Meeting of the Association for Computational
  Linguistics (ACL 2016). pp. 409--420 (2016)

\bibitem{Cai2017Fast}
Cai, D., Zhao, H., Zhang, Z., Xin, Y., Wu, Y., Huang, F.: Fast and accurate
  neural word segmentation for {Chinese}. In: Proceedings of the 55th Annual
  Meeting of the Association for Computational Linguistics (ACL 2017). pp.
  608--615 (2017)

\bibitem{Cho2014Learning}
Cho, K., Merrienboer, B.V., Gulcehre, C., Bahdanau, D., Bougares, F., Schwenk,
  H., Bengio, Y.: Learning phrase representations using rnn encoder-decoder for
  statistical machine translation. In: Proceedings of the 2014 Conference on
  Empirical Methods in Natural Language Processing (EMNLP 2014). pp. 1724--1734
  (2014)

\bibitem{Cui2016Attention}
Cui, Y., Chen, Z., Wei, S., Wang, S., Liu, T., Hu, G.: Attention-over-attention
  neural networks for reading comprehension. In: Proceedings of the 55th Annual
  Meeting of the Association for Computational Linguistics (ACL 2017). pp.
  1832--1846 (2017)

\bibitem{Cui2017Dataset}
Cui, Y., Liu, T., Chen, Z., Ma, W., Wang, S., Hu, G.: Dataset for the first
  evaluation on chinese machine reading comprehension. In: chair), N.C.C.,
  Choukri, K., Cieri, C., Declerck, T., Goggi, S., Hasida, K., Isahara, H.,
  Maegaard, B., Mariani, J., Mazo, H., Moreno, A., Odijk, J., Piperidis, S.,
  Tokunaga, T. (eds.) Proceedings of the Eleventh International Conference on
  Language Resources and Evaluation (LREC 2018). European Language Resources
  Association (ELRA) (2018)

\bibitem{Cui2016Consensus}
Cui, Y., Liu, T., Chen, Z., Wang, S., Hu, G.: Consensus attention-based neural
  networks for {Chinese} reading comprehension. In: booktitle={Proceedings of
  the 26th International Conference on Computational Linguistics (COLING
  2016)}. pp. 1777--1786 (2016)

\bibitem{Dhingra2017Gated}
Dhingra, B., Liu, H., Yang, Z., Cohen, W.W., Salakhutdinov, R.: Gated-attention
  readers for text comprehension. In: Proceedings of the 55th Annual Meeting of
  the Association for Computational Linguistics (ACL 2017). pp. 1832--1846
  (2017)

\bibitem{He2018Syntax}
He, S., Li, Z., Zhao, H., Bai, H., Liu, G.: Syntax for semantic role labeling,
  to be, or not to be. In: Proceedings of the 56th Annual Meeting of the
  Association for Computational Linguistics (ACL 2018) (2018)

\bibitem{hermann2015teaching}
Hermann, K.M., Kocisky, T., Grefenstette, E., Espeholt, L., Kay, W., Suleyman,
  M., Blunsom, P.: Teaching machines to read and comprehend. In: Advances in
  Neural Information Processing Systems (NIPS 2015). pp. 1693--1701 (2015)

\bibitem{hill2015goldilocks}
Hill, F., Bordes, A., Chopra, S., Weston, J.: The goldilocks principle: Reading
  children's books with explicit memory representations. arXiv preprint
  arXiv:1511.02301  (2015)

\bibitem{Huang2018Moon}
Huang, Y., Li, Z., Zhang, Z., Zhao, H.: {Moon IME:} neural-based chinese pinyin
  aided input method with customizable association. In: Proceedings of the 56th
  Annual Meeting of the Association for Computational Linguistics (ACL 2018),
  System Demonstration (2018)

\bibitem{Joshi2017TriviaQA}
Joshi, M., Choi, E., Weld, D.S., Zettlemoyer, L.: Triviaqa: A large scale
  distantly supervised challenge dataset for reading comprehension. In: ACL.
  pp. 1601--1611 (2017)

\bibitem{kadlec2016text}
Kadlec, R., Schmid, M., Bajgar, O., Kleindienst, J.: Text understanding with
  the attention sum reader network. In: Proceedings of the 54th Annual Meeting
  of the Association for Computational Linguistics (ACL 2016). pp. 908--918
  (2016)

\bibitem{Kim2015Character}
Kim, Y., Jernite, Y., Sontag, D., Rush, A.M.: Character-aware neural language
  models. In: Proceedings of the Thirtieth AAAI Conference on Artificial
  Intelligence (AAAI 2016). pp. 2741--2749 (2016)

\bibitem{kingma2014adam}
Kingma, D., Ba, J.: Adam: A method for stochastic optimization. arXiv preprint
  arXiv:1412.6980  (2014)

\bibitem{lample2016neural}
Lample, G., Ballesteros, M., Subramanian, S., Kawakami, K., Dyer, C.: Neural
  architectures for named entity recognition. arXiv preprint arXiv:1603.01360
  (2016)

\bibitem{Ling2015Character}
Ling, W., Trancoso, I., Dyer, C., Black, A.W.: Character-based neural machine
  translation. arXiv preprint arXiv:1511.04586  (2015)

\bibitem{luong2016achieving}
Luong, M.T., Manning, C.D.: Achieving open vocabulary neural machine
  translation with hybrid word-character models. arXiv preprint
  arXiv:1604.00788  (2016)

\bibitem{mikolov:2013}
Mikolov, T., Chen, K., Corrado, G., Dean, J.: Efficient estimation of word
  representations in vector space. arXiv preprint arXiv:1301.3781  (2013)

\bibitem{Miyamoto2016Gated}
Miyamoto, Y., Cho, K.: Gated word-character recurrent language model. In:
  Proceedings of the 2016 Conference on Empirical Methods in Natural Language
  Processing (EMNLP 2016). pp. 1992--1997 (2016)

\bibitem{Munkhdalai2016Reasoning}
Munkhdalai, T., Yu, H.: Reasoning with memory augmented neural networks for
  language comprehension. Proceedings of the International Conference on
  Learning Representations (ICLR 2017)  (2017)

\bibitem{Peters2018Deep}
Peters, M.E., Neumann, M., Iyyer, M., Gardner, M., Clark, C., Lee, K.,
  Zettlemoyer, L.: Deep contextualized word representations. In: Conference of
  the North American Chapter of the Association for Computational Linguistics:
  Human Language Technologies (NAACL 2018) (2018)

\bibitem{rajpurkar2016squad}
Rajpurkar, P., Zhang, J., Lopyrev, K., Liang, P.: {SQuAD}: 100,000+ questions
  for machine comprehension of text. In: Proceedings of the 2016 Conference on
  Empirical Methods in Natural Language Processing (EMNLP 2016). pp. 2383--2392
  (2016)

\bibitem{Seo2016Bidirectional}
Seo, M., Kembhavi, A., Farhadi, A., Hajishirzi, H.: Bidirectional attention
  flow for machine comprehension. In: Proceedings of the International
  Conference on Learning Representations (ICLR 2017) (2017)

\bibitem{sordoni2016iterative}
Sordoni, A., Bachman, P., Trischler, A., Bengio, Y.: Iterative alternating
  neural attention for machine reading. arXiv preprint arXiv:1606.02245  (2016)

\bibitem{Trischler2016Natural}
Trischler, A., Ye, Z., Yuan, X., Suleman, K.: Natural language comprehension
  with the epireader. In: Proceedings of the 2016 Conference on Empirical
  Methods in Natural Language Processing (EMNLP 2016). pp. 128--137 (2016)

\bibitem{Wang2017Conditional}
Wang, B., Liu, K., Zhao, J.: Conditional generative adversarial networks for
  commonsense machine comprehension. In: Proceedings of the Twenty-Sixth
  International Joint Conference on Artificial Intelligence (IJCAI 2017). pp.
  4123--4129 (2017)

\bibitem{Wang2016Machine}
Wang, S., Jiang, J.: Machine comprehension using {Match-LSTM} and answer
  pointer. Proceedings of the International Conference on Learning
  Representations (ICLR 2016)  (2016)

\bibitem{Wang2017Gated}
Wang, W., Yang, N., Wei, F., Chang, B., Zhou, M.: Gated self-matching networks
  for reading comprehension and question answering. In: Proceedings of the 55th
  Annual Meeting of the Association for Computational Linguistics (ACL 2017).
  pp. 189--198 (2017)

\bibitem{Yizhong2018Multi}
Wang, Y., Liu, K., Liu, J., He, W., Lyu, Y., Wu, H., Li, S., Wang, H.:
  Multi-passage machine reading comprehension with cross-passage answer
  verification. In: Proceedings of the 56th Annual Meeting of the Association
  for Computational Linguistics (ACL 2018) (2018)

\bibitem{Yang2016Words}
Yang, Z., Dhingra, B., Yuan, Y., Hu, J., Cohen, W.W., Salakhutdinov, R.: Words
  or characters? fine-grained gating for reading comprehension. In: Proceedings
  of the International Conference on Learning Representations (ICLR 2017)
  (2017)

\bibitem{zhang2018SubMRC}
Zhang, Z., Huang, Y., Zhao, H.: Subword-augmented embedding for cloze reading
  comprehension. In: Proceedings of the 27th International Conference on
  Computational Linguistics (COLING 2018) (2018)

\bibitem{zhang2018DUA}
Zhang, Z., Li, J., Zhu, P., Zhao, H.: Modeling multi-turn conversation with
  deep utterance aggregation. In: Proceedings of the 27th International
  Conference on Computational Linguistics (COLING 2018) (2018)

\bibitem{zhang2018OneShot}
Zhang, Z., Zhao, H.: One-shot learning for question-answering in gaokao history
  challenge. In: Proceedings of the 27th International Conference on
  Computational Linguistics (COLING 2018) (2018)

\bibitem{Zhu2018lingke}
Zhu, P., Zhang, Z., Li, J., Huang, Y., Zhao, H.: Lingke: A fine-grained
  multi-turn chatbot for customer service. In: Proceedings of the 27th
  International Conference on Computational Linguistics (COLING 2018), System
  Demonstrations (2018)

\end{thebibliography}
\end{document}